\documentclass[letterpaper]{article} % DO NOT CHANGE THIS
\usepackage{aaai23}  % DO NOT CHANGE THIS
\usepackage{times}  % DO NOT CHANGE THIS
\usepackage{helvet}  % DO NOT CHANGE THIS
\usepackage{courier}  % DO NOT CHANGE THIS
\usepackage[hyphens]{url}  % DO NOT CHANGE THIS
\usepackage{graphicx} % DO NOT CHANGE THIS
\usepackage{natbib}  % DO NOT CHANGE THIS AND DO NOT ADD ANY OPTIONS TO IT
\usepackage{caption} % DO NOT CHANGE THIS AND DO NOT ADD ANY OPTIONS TO IT
\usepackage{booktabs}
\usepackage{algorithm}
\usepackage{algorithmic}

\usepackage{newfloat}
\usepackage{listings}
\DeclareCaptionStyle{ruled}{labelfont=normalfont,labelsep=colon,strut=off} % DO NOT CHANGE THIS
\floatstyle{ruled}
\newfloat{listing}{tb}{lst}{}
\floatname{listing}{Listing}
\usepackage{color }
\usepackage{subfigure}
\usepackage{multirow}
\usepackage{amsfonts} 
\usepackage{soul}
\title{Instance Smoothed Contrastive Learning for Unsupervised Sentence Embedding}

\author{
    Hongliang He\textsuperscript{\rm 1,2}\equalcontrib,
    Junlei Zhang\textsuperscript{\rm 1,2}\equalcontrib,
    Zhenzhong Lan\textsuperscript{\rm 2,3}\thanks{Corresponding authors.}, 
    Yue Zhang\textsuperscript{\rm 2,3}\footnotemark[2]
}
\affiliations{
    \textsuperscript{\rm 1}Zhejiang University, China\\
    \textsuperscript{\rm 2}School of Engineering, Westlake University, China\\
    \textsuperscript{\rm 3}Institute of Advanced Technology, Westlake Institute for Advanced Study, China \\
     \{hehongliang, zhangjunlei, lanzhenzhong, zhangyue\}@westlake.edu.cn

}

\begin{document}

\maketitle

\begin{abstract}
Contrastive learning-based methods, such as unsup-SimCSE, have achieved state-of-the-art (SOTA) performances in learning unsupervised sentence embeddings.
However, in previous studies, each embedding used for contrastive learning only derived from one sentence instance, and we call these embeddings \textbf{instance-level} embeddings. In other words, each embedding is regarded as a unique class of its own, which may hurt the generalization performance. In this study, we propose IS-CSE (\underline{i}nstance \underline{s}moothing \underline{c}ontrastive \underline{s}entence \underline{e}mbedding) to smooth the boundaries of embeddings in the feature space. Specifically, we retrieve embeddings from a dynamic memory buffer according to the semantic similarity to get a positive embedding group. Then embeddings in the group are aggregated by a self-attention operation to produce a \textbf{smoothed instance} embedding for further analysis. We evaluate our method on standard semantic text similarity (STS) tasks and achieve an average of  $78.30\%$,  $79.47\%$,  $77.73\%$, and  $79.42\%$ Spearman's correlation on the base of BERT-base, BERT-large, RoBERTa-base, and RoBERTa-large respectively, a $2.05\%$,  $1.06\%$,  $1.16\%$ and  $0.52\%$ improvement compared to unsup-SimCSE. 
\end{abstract}

\section{Introduction}
Learning better universal sentence embedding \cite{gao2021simcse} can benefit many natural language processing tasks, such as sentiment analysis, information retrieval and semantic search \cite{klein2022micse,zhang2018illustrate,pilehvar2015senses}, and thus has received much attention. Recently, it has been shown that the contrastive learning-based methods give strong results for sentence embeddings
\cite{gao2021simcse,wang2022improving,zhou2022debiased,zhang2021s}. The core idea of contrastive learning is that positive and negative embedding pairs are generated given a batch of training sentences. Whereas the positive embeddings are often obtained via augmentation, and negative embeddings are sampled from a random collection of sentences. Following the construction of pairs,  contrastive learning forces the model to learn discriminative embeddings by pulling positive sentence pairs together and pushing apart negative ones.

In the unsupervised contrastive learning framework, while some works seek to optimize for selecting "hard" negative examples \cite{zhou2022debiased} or using pre-defined prompt \cite{jiang2022promptbert} to extract features, other methods investigate the effects of augmentation on constructing sentence pairs. One of the most influential methods for learning sentence embeddings is SimCSE \cite{gao2021simcse}, which takes drop-out as data augmentation, providing expressive semantically similar embeddings to construct positive pairs. ESimCSE\cite{wu2021esimcse} augmented the input sentences by word repetition, insertion, and deletion. Similarly, CARDS \cite{wang2022improving} randomly flip the first letter in a word to augment the inputs.

\begin{figure}[t!]
  \includegraphics[width=1.01\linewidth]{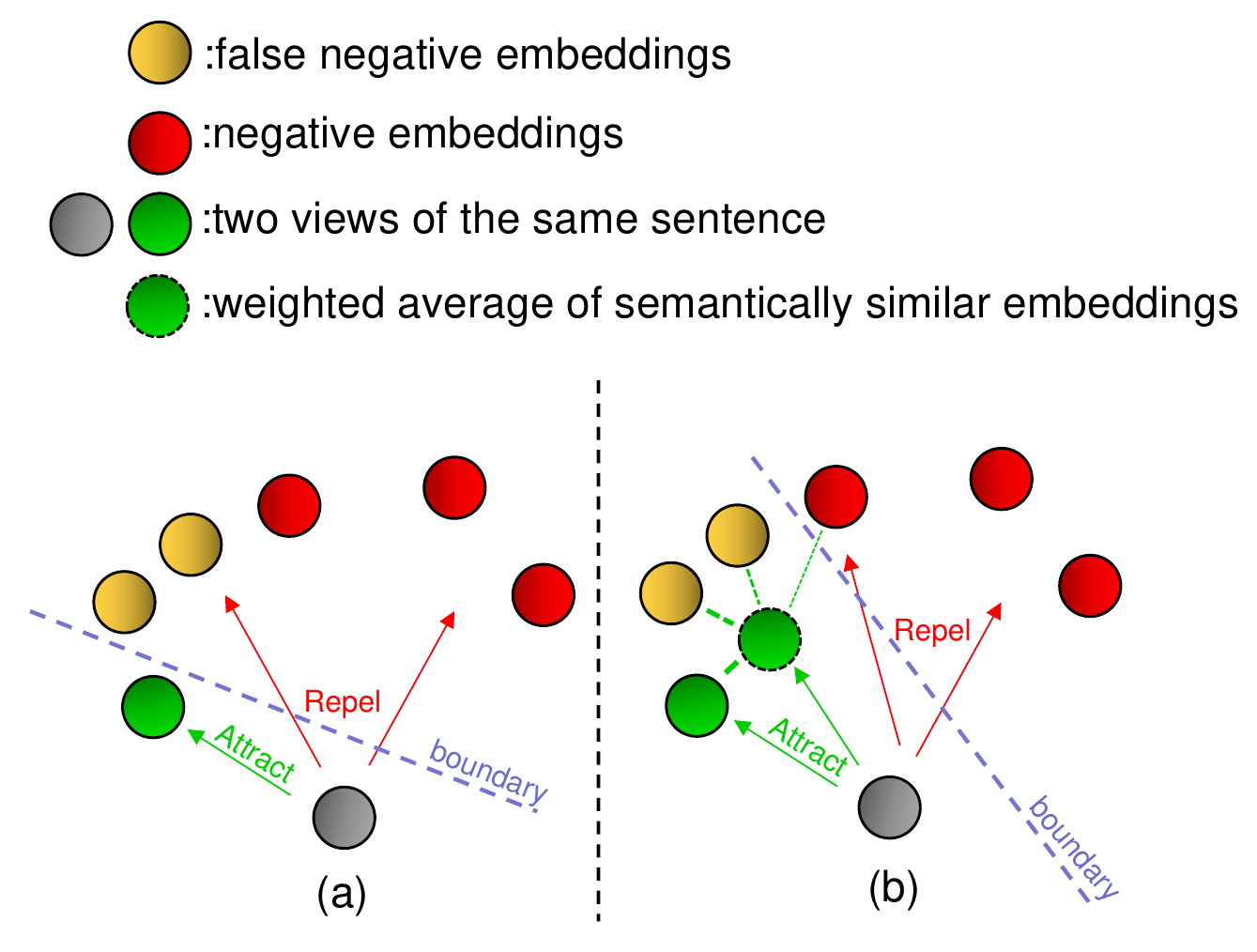}
  \caption{Comparison between our method with SimCSE. In SimCSE, two views of the same input sentence are regarded as positive pairs. Other sentences in the same batch are regarded as negative examples. (a): In SimCSE, each embedding is derived from one sentence, and one view of the input sentence is regarded as a label of another view. (b): Our method uses additional soft labels (weighted average of closing-by embeddings).}
  \label{fig: simcsevsiscse1}
\end{figure}

However, most of these methods take each of the sentences as a unique class and discriminate it from other sentences in a batch. This could make models become "over-confident"  about each sentence being a separate class, because there may be some false negative pairs in an unsupervised setting. To address this problem, DCLR \cite{zhou2022debiased} generates negative examples by sampling them from a learned gaussian distribution and filtering out negative examples with high similarities. However, DCLR does not make use of rich positive embeddings.  Inspired by the success of label smoothing \cite{muller2019does} where soft labels are applied to release the "over-confident" of a network caused by hard labels, we propose to smooth the positive examples to release the "over-confident" problem. For the positive pairs in contrastive learning, one positive embedding can be regarded as a label which another positive one should fit. Following the label smoothing method, we smooth the label by a weighted average operation with retrieved semantically similar embeddings. Specifically, we hold a First-in-First-out memory buffer which saves the sentence embeddings in the previous steps during the training process. While constructing the positive pairs,  we retrieve sentence embeddings from the memory buffer based on the cosine similarity and do a weighted average operation with the positive embedding to get smooth embeddings. This can push each sentence to be similar to other closing-by sentences, not just itself. This new practice has a label smoothing effect \cite{szegedy2016rethinking}. We call it instance smoothing to contrast sentence embedding (IS-CSE).

We evaluate IS-CSE on seven standard semantic textual similarity (STS) tasks \cite{agirre2012semeval,agirre2013sem,agirre2014semeval,agirre2015semeval,agirre2016semeval,cer2017semeval,marelli2014sick} and 7 transfer learning tasks \cite{pang2005seeing,hu2004mining,wiebe2005annotating,socher2013recursive,voorhees2000building,dolan2005automatically}. Results show that our unsupervised model achieves a $79.47\%$ and $79.42\%$ averaged Spearman's correlation respectively using BERT$_{large}$ and RoBERTa$_{large}$, significantly outperforming competitive baselines on STS tasks. To better understand the effect of group-level embedding, we also calculate the alignment score \cite{wang2020understanding} between semantically similar positive pairs and the uniformity score of the whole representation to measure the quality of learned embeddings. We find that IS-CSE achieves better alignment results. But there is a little drop in uniformity except for the BERT$_{large}$ model. To the best of our knowledge, IS-CSE is the first attempt to create positive pairs from a group of similar sentences rather than each sentence in contrastive learning of unsupervised sentence representations. Our code is available at \url{https://github.com/dll-wu/IS-CSE}
% However, there are some adverse effects on uniformity. But we argue that the uniformity only drops 0.04 on RoBERTa$_{large}$ and even increases 0.31 on BERT$_{large}$.
% We find that our IS-CSE  improves alignment with a slight decrease in alignment, which is a better balance between alignment and uniformity. 

% Our contributions are summarized as follows:

% \begin{enumerate}

%  \item  To the best of our knowledge, IS-CSE is the first attempt to create positive pairs from a group of similar sentences rather than each sentence in contrastive learning of unsupervised sentence representations.  

%  \item We proposed a dynamic memory buffer for efficient similar sentence retrieval in IS-CSE training. 

%  \item Experimental results on seven semantic textual similarity tasks show the effectiveness of our methods.

% \end{enumerate}

\section{Related Work}

\subsection{Unsupervised Sentence Embedding Learning}
Unsup-SimCSE \cite{gao2021simcse} proposes a contrastive learning framework to finetune pre-trained BERT \cite{devlin2018bert} and RoBERTa \cite{liu2019roberta}, and significantly outperforms previous results. SimCSE is further enhanced by several follow-up studies from different prospects. Instead of simply representing the sentence with [CLS] token or averaged embeddings, PromptBERT \cite{jiang2022promptbert} uses prompt tokens to represent a sentence. Data augmentation methods \cite{wu2021esimcse,yan2021consert} are also applied to produce more high-quality training samples. For example, ESimCSE \cite{wu2021esimcse} enhances the input sentences with a repetition operation; ConSERT \cite{yan2021consert} takes multiple data augmentation strategies to further generate views for contrastive learning. Besides data augmentation methods, DCLR \cite{zhou2022debiased} proposes an instance weighting method to punish false negatives and generate noise-based negatives to guarantee the uniformity of the representation space.  However, no previous work in this task has tried to smooth the positive instances with sampled semantically similar sentence embeddings. Our work mainly differs from previous unsupervised embedding learning methods in three prospects: 1) we use a dynamic buffer to reduce the computational consumption; 2) we aggregate the retrieved embeddings to form a smoothed embedding instead of using their instance-level embedding directly in previous works;  3) we use both the instance and smoothed instance embeddings for discrimination.

\subsection{Contrastive Learning}

Contrastive learning has been originated applied in computer vision \cite{he2020momentum,chen2020improved} and information retrieval \cite{bian2021contrastive} and achieved significant performance improvement. Data augmentation strategies such as image rotation and random cropping \cite{gao2021simcse,bian2021contrastive,li2020residual} are used to produce augmented images. The augmented images are then used as positive images for discrimination, while other images in the same mini-batch are regarded as negative ones. For unsupervised sentence representation learning, SimCSE \cite{gao2021simcse} adopts dropout as the data augmentation, which improves the results on semantic textual similarities tasks by a large margin. Subsequent studies further adopt token shuffling \cite{yan2021consert} and back translation \cite{fang2020cert} to augment positive examples for sentence representation learning. However, to the best of our knowledge, how to augment embeddings from the smoothing view has not been studied. We fill the gap by investigating the effect of embedding smoothing for unsupervised sentence embedding learning.

\begin{figure}[t!]
  \includegraphics[width=1.01\linewidth]{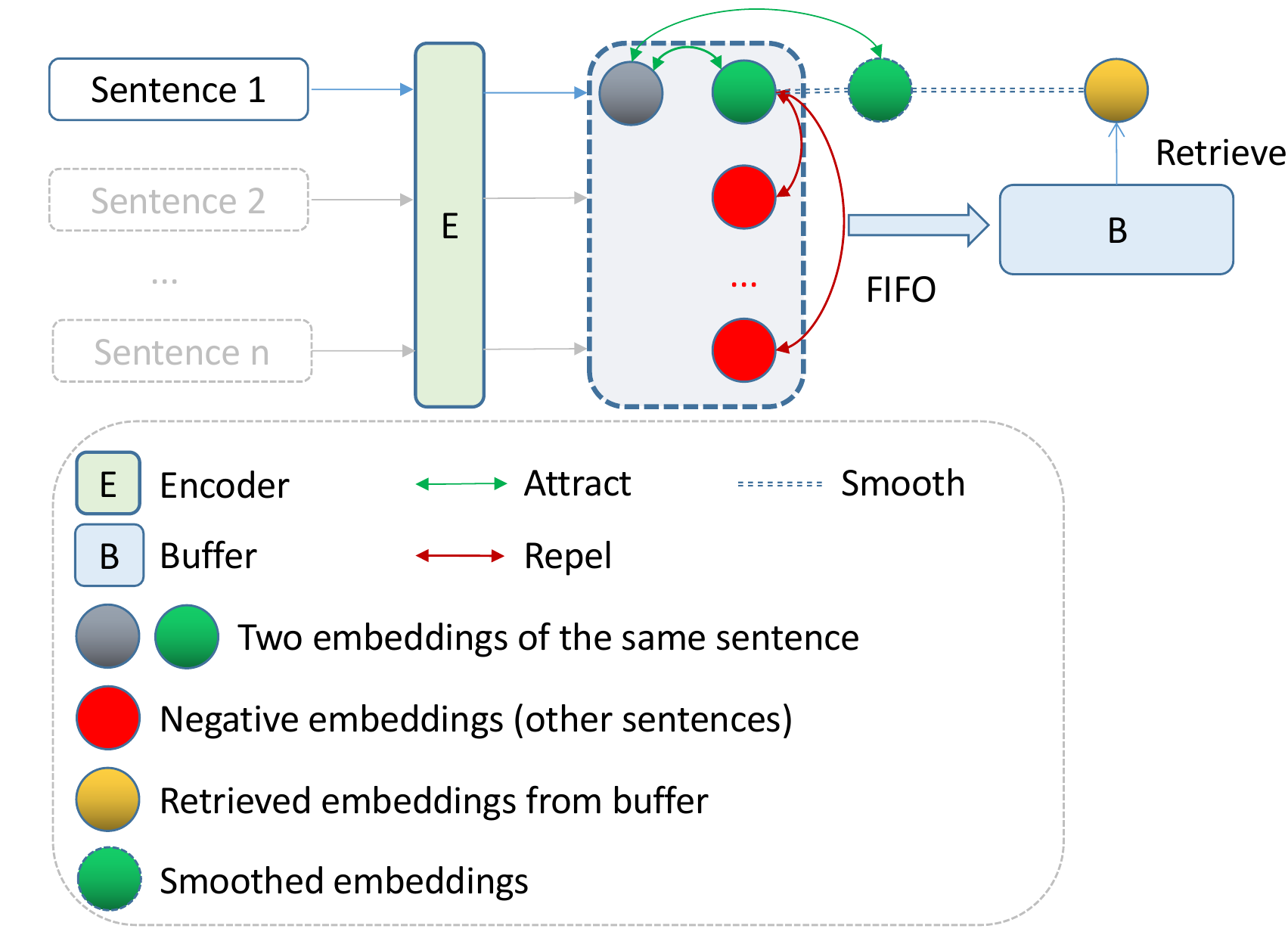}
  \caption{Overview of our method. We retrieve embeddings from the memory buffer (orange) and the smoothed embeddings are the weighted average of the retrieved and positive embeddings.}
  \label{fig: simcsevsiscse}
\end{figure}

\section{Method}

\subsection{Baseline}

SimCSE \cite{gao2021simcse} applies contrastive learning on the universal sentence learning problem, where instance-level sentence embeddings are used as the input of the InfoNCE loss \cite{oord2018representation}. Specifically, given a collection of input sentences $\{x_i\}_{i=1}^m$. SimCSE simply uses identical sentences to build the sentence pair, i.e. $x_i = x_i^{+}$, and feeds $x_i$ into a Transformer encoder $f_\theta$ twice. Since independently sampled dropout masks are applied to fully-connected layers and attention probabilities in $f_\theta$, two separate sentence embeddings $h_i$ and $h_i^{+}$ are obtained. For a mini-batch of $N$ samples, the training loss for unsupervised SimCSE (unsup-SimCSE): %Specifically, given a pair of input sentences $\left\{x_i, x_i^{+}\right\}$, where $x_i$ and $x_i^{+}$ is a semantically related sentence pair and $i$ is the index of the sentences in a mini-batch. let $h_i$ and $h_i^{+}$ denote the projected instance-level embeddings of $x_i$ and $x_i^{+}$ by a shared encoder $f_\theta$. 
\begin{equation}
    \label{equ:simclrloss}
    \mathcal{L}_{instance} =-log\frac{e^{sim(h_i, h_i^{+}) / \tau} }{\sum_{j=1}^{N} e^{sim(h_i, h_j^+) / \tau}},
\end{equation}
where $\tau$ is a temperature parameter and the $sim(\cdot, \cdot)$ represents the cosine similarity function:
\begin{equation}
    \label{equ:cosinesim}
    sim(h_i, h_i^{+})=\frac{h_i^{T}h_i^{+}}{\left \| h_i \right \|\left \| h_i^{+} \right \|}.
\end{equation}

All embeddings $h$ in Equ.\ref{equ:simclrloss} are instance-level embeddings, each of which is derived from one sentence instance. In this paper, we propose an instance-smoothing mechanism to regularize the InfoNCE loss by applying smoothed instance embeddings (derived from a group of semantically similar sentences). 

%To reduce the computational consumption of retrieving sentences, we deploy a dynamic memory bank to store the embeddings computed in previous batches. To the best of our knowledge, this paper is the first to study the instance-smoothing mechanism with a memory buffer in unsupervised sentence representation learning.

\subsection{Dynamic Memory Buffer}
Instead of only using embeddings derived from input sentences, we construct smoothed embeddings by averaging the closing-by embeddings. One key process of IS-CSE is to retrieve these closing-by embeddings at each step during finetuning. Directly retrieving sentence embeddings from the whole dataset can lead to a huge computational burden. To bound the memory usage, we propose to use a dynamic memory buffer in the unsupervised contrastive learning task. Specifically, given a dynamic memory buffer $\mathcal{B} \in 
 \mathbb{R}^{L \times d}$, where $L$ is the length of the buffer and $d$ is the dimension of an embedding. For each step, we feed the buffer normalized augmented embeddings $h^+$ with a First-in-first-out (FIFO) strategy. The embeddings in memory buffer $\mathcal{B}$ are stop-gradient embeddings. Formally, the method for updating the memory buffer $\mathcal{B}$ is:
 \begin{equation}
     \mathcal{B}_{new} = Concat(\mathcal{B}_{old}[l:L], sg\{\frac{h_1^+}{||h_1^+||},...,\frac{h_{l}^+}{||h_{l}^+||} \} ),
 \end{equation}
where $l$ is the number of coming/discarded embeddings for the FIFO strategy ($l$ equals the batch size in our experiment), $sg$ is the stop-gradient operation and $Concat$ operation is used to maintain the buffer size and dynamically update the buffer. Based on the memory buffer, several semantically similar embeddings are retrieved for smoothing the augmented positive embedding $h^+$. 

\subsection{Retrieving Sentence Embeddings}
After setting up the dynamic memory buffer, we retrieve sentence representations and apply the weighted average operation to get the smoothed embeddings. We compare two types of retrieval methods: kNN and K-means.
\subsubsection{kNN} A simple way to obtain semantically similar embeddings is kNN \cite{peterson2009k}. Given the augmented embedding $h^+$, we calculate the cosine similarly (Equ.\ref{equ:cosinesim}) between $h^+$ and each of the embedding in buffer $\mathcal{B}$. Then $k$ nearest embeddings are retrieved from $\mathcal{B}$.

\subsubsection{K-means} We perform the K-means algorithm \cite{hartigan1979algorithm} on $\mathcal{B}$ with a pre-defined number of clusters $k'$. We assign each embedding to a cluster based on semantic similarity. We directly retrieve the center embedding to which $h^+$ belongs.

We empirically compare the performances of kNN and K-means in Table \ref{tab:knn_and_kmeans} and select the kNN as our final retrieval method.

\subsection{Smoothing Instance Embeddings} In IS-CSE, the augmented embeddings $h^+$ are smoothed by retrieved embeddings with high semantic similarity from the dynamic buffer. For kNN, we apply a self-attention aggregation method. Specifically, given $k$ retrieved embeddings $\left\{ h^r \right\}_{i=1}^{k}$ and the augmented embedding $h^+$, we normalized and then concatenate them to get a combined matrix $K = \{h^+, h^r_1, h^r_2, ...,h^r_k\} \in \mathbb{R}^{(k+1)\times d}$. We thus obtain smoothed embedding $h^{s+}$ by:

\begin{equation}
    \label{equ:selfattnaggr}
    h^{s+} = softmax(\frac{h^{+} K^T}{\beta}) K ,
\end{equation}
where $\beta$ is a temperature parameter. 

For K-means, we cluster all the embeddings in the buffer based on the cosine similarity. Then we obtain a list of cluster centers, and select the center $c^+$ of the cluster which $h^+$ belongs to. We get our smoothed embedding $h^{s+}$ by:

\begin{equation}
    \label{equ:weightedaggr}
    h^{s+} = \gamma h^{+} + (1 - \gamma) c^+ ,
\end{equation}
where $\gamma$ is a hyper-parameter.
In Equ.\ref{equ:selfattnaggr} and Equ.\ref{equ:weightedaggr}, $h^+$ is not a  stop-gradient embedding but the retrieved embeddings $h^r$ and centers $c^+$ are stop-gradient embeddings.

\begin{table*}[t!]

\centering
\begin{tabular}{@{}lllllllll@{}}
\toprule
Model                            & STS12 & STS13 & STS14 & STS15 & STS16 & STS-B & SICK-R & Avg.  \\ \midrule
GloVe embeddings(avg.)$^*$           & 55.14 & 70.66 & 59.73 & 68.25 & 63.66 & 58.02 & 53.76  & 61.32 \\
BERT$_{base}$(first-last avg.)$^*$     & 39.70 & 59.38 & 49.67 & 66.03 & 66.19 & 53.87 & 62.06  & 56.80 \\
BERT$_{base}$-flow$^*$             & 58.40 & 67.10 & 60.85 & 75.16 & 71.22 & 68.66 & 64.47  & 66.55 \\
BERT$_{base}$-whitening$^*$            & 57.83 & 66.90 & 60.90 & 75.08 & 71.31 & 68.24 & 63.74  & 66.28 \\
IS-BERT$_{base}^*$                   & 56.77 & 69.24 & 61.21 & 75.23 & 70.16 & 69.21 & 64.25  & 66.58 \\
CT-BERT$_{base}^*$                   & 61.63 & 76.80 & 68.47 & 77.50 & 76.48 & 74.31 & 69.19  & 72.05 \\
SimCSE-BERT$_{base}^*$             & 68.40 & 82.41 & 74.38 & 80.91 & 78.56 & 76.85 & 72.23  & 76.25 \\ 
DLCR-BERT$_{base}^*$    & \underline{70.81} & \underline{83.73} & \underline{75.11} & \underline{82.56} & \underline{78.44} & \underline{78.31} & \underline{71.59}  & \underline{77.22} \\ 
IS-CSE-BERT$_{base}$             & \textbf{72.86} & \textbf{84.02} & \textbf{76.35} & \textbf{82.64} & \textbf{78.65} & \textbf{79.53} & \textbf{74.05} & \textbf{78.30}  \\
\midrule
SimCSE-BERT$_{large}^*$             & 70.88 & 84.16 & 76.43 & 84.50 & \underline{79.76} & 79.26 & 73.88   & 78.41 \\ 
 DCLR-BERT$_{large}$          & \underline{71.87} & \underline{84.83} & \underline{77.37} & \underline{84.70} & \textbf{79.81} & \underline{79.55} & \underline{74.19} & \underline{78.90 }   \\
IS-CSE-BERT$_{large}$            & \textbf{73.76}   & \textbf{85.06} & \textbf{78.14} & \textbf{85.02} & 79.59 & \textbf{80.43} & \textbf{74.30} & \textbf{79.47}  \\
\midrule
RoBERTa$_{base}$(fist-last avg.)$^*$   & 40.88 & 58.74 & 49.07 & 65.63 & 61.48 & 58.55 & 61.63  & 56.57 \\
RoBERTa$_{base}$-whitening$^*$         & 46.99 & 63.24 & 57.23 & 71.36 & 68.99 & 61.36 & 62.91  & 61.73 \\
DeCLUTR-RoBERTa$_{base}^*$         & 52.41 & 75.19 & 65.52 & 77.12 & 78.63 & 72.41 & 68.62  & 69.99 \\
SimCSE-RoBERTa$_{base}^*$          & \underline{70.16} & 81.77 & 73.24 & 81.36 & 80.65 & 80.22 & 68.56  & 76.57 \\
% CARDS-RoBERTa$_{base}$          &  &  &  & &  &  & &  \\
% CARDS-RoBERTa$_{large}$          &  &  &    &   &  &  &   &  \\
 DCLR-RoBERTa$_{base}$          & 70.01 & \textbf{83.08} & \textbf{75.09} & \textbf{83.66} & 81.06 & \textbf{81.86} & \textbf{70.33} & \textbf{77.87}  \\ 
IS-CSE-RoBERTa$_{base}$       & \textbf{71.39} & \underline{82.58}&	\underline{74.36} &	\underline{82.75} &	\textbf{81.61} & \underline{81.40}	 & \underline{69.99} &	\underline{77.73} \\  \midrule
SimCSE-RoBERTa$_{large}^*$         & 72.86 & 83.99 & 75.62 & 84.77 & 81.80 & 81.98 & 71.26  & 78.90 \\
% DCLR-RoBERTa$_{large}$          & \textbf{73.09} & 84.57 & 76.13 & 85.15 & \textbf{81.99} & 82.35 & \textbf{71.80} & 79.30  \\
DCLR-RoBERTa$_{large}^{\dagger}$          & 73.09 & 84.57 & 76.13 & 85.15 & 81.99 & 82.35 & 71.80 & 79.30 \\
DCLR-RoBERTa$_{large}$ (ours)           & 71.30 & 84.67 & 76.17 & 84.65 & 81.62 &  81.93   & \underline{72.29} & 78.95   \\
CARDS-RoBERT$_{large}$ (ours)           & \textbf{74.78} & \underline{86.42} & \underline{79.02} & \underline{85.95} & \underline{82.36} & \underline{83.65}   & 70.81 &  \underline{80.46}   \\
IS-CSE-RoBERTa$_{large}$          & 72.84 & 85.02 & 76.99 & 85.58 & 80.93 & 82.87 & 71.68  & 79.42  \\
% \quad + DCLR & 72.84 & \textbf{85.02} & \textbf{76.99} & \textbf{85.58}& 80.93 & \textbf{82.87} & 71.68  & \textbf{79.42} \\ 

\quad + DCLR     & 73.67 & 85.46 & 76.86 & 85.16 & 81.31 & 82.25     & 71.71 & 79.49 \\

\quad + CARDS     & \underline{74.30} & \textbf{86.47} & \textbf{79.06} & \textbf{85.99} & \textbf{82.78} & \textbf{84.02}     & \textbf{72.80} & \textbf{80.77} \\

\bottomrule
\end{tabular}
\caption{Sentence embedding performance on STS tasks (Spearman's correlation). The best performance and the second-best performance with the same pre-trained encoder are denoted in bold and underlined fonts respectively. $^*$: results from \cite{gao2021simcse}; $^{\dagger}$: results from \cite{zhou2022debiased}; 
(ours): our reproduced results based on code released by their authors; We add our $L_{smoothing}$ to the DCLR to get combined results and show it on "+DCLR". All the experiments are conducted in an unsupervised setting.}

  \label{tabel: sts_main_resutls}

\end{table*}

\subsection{Instance Smoothing Contrastive Sentence
Embedding (IS-CSE)} 
The main difference between our method and SimCSE is that we add an additional contrastive loss whose augmented positive embeddings are smoothed. Given a batch of input sentences, we obtain the projected instance-level embeddings of $h_i$ and $h_i^+$. We calculate our smoothed embedding $h_i^{s+}$ using Equ.\ref{equ:selfattnaggr}. The smoothed embedding loss can be calculated by:
\begin{equation}
    \label{equ:smoothing}
    \mathcal{L}_{smoothing} =-log\frac{e^{sim(h_i, h_i^{s+}) / \tau} }{\sum_{j=1}^{N} e^{sim(h_i, h_j^{s+}) / \tau}} .
\end{equation}

Combining Equ.\ref{equ:simclrloss} and Equ.\ref{equ:smoothing}, we treat the smoothing loss as a regularizer. The final form of our training objective is:
\begin{equation}
    \label{equ:finalloss}
    \mathcal{L} = \mathcal{L}_{instance} + \alpha \mathcal{L}_{smoothing} ,
\end{equation}
where $\alpha$ is a coefficient.

The quality of retrieved embeddings may be low at the initial stages because the model has not been fully finetuned. A big $\alpha$ may hurt the model performance at the initial stages of finetuning. We adopt a cosine scheduler for $\alpha$:
\begin{equation}
\label{equ: cosine_alpha}
    \alpha = \min\{\cos(\pi \cdot \frac{T_{i}}{T_{max}}) * (\alpha_{start}-\alpha_{end}) , 0\}+ \alpha_{end} ,
\end{equation}
where $\alpha_{start}$ , $\alpha_{end}$, $T_{i}$ and $T_{max}$ are the inital value of $\alpha$, end value of $\alpha$, the current step and the max step, respectively.
%%%%%%%%%%%%%%%%%%%%%%%%%%%%%Experiments%%%%%%%%%%%%%%%%%%%%%%%%%%%%%%%%%%%%%%%%%%%%%%%
%%%%%%%%%%%%%%%%%%%%%%%%%%%%%%%%%%%%%%%%%%%%%%%%%%%%%%%%%%%%%%%%%%%%%%%%%%%%%%%%%%%%%%%

\section{Experiments}
\subsection{Setup}
 For unsupervised sentence embedding learning, we follow the same training process as SimCSE \cite{gao2021simcse}.  We conduct our main experiments on 7 standard semantic textual similarities (STS) tasks: STS 2012-2016 \cite{agirre2012semeval,agirre2013sem,agirre2014semeval,agirre2015semeval,agirre2016semeval}, STS Benchmark \cite{cer2017semeval} and SICK-Relatedness \cite{marelli2014sick}. We compare our IS-CSE against methods reported in SimCSE \cite{gao2021simcse} and SimCSE-related methods: DCLR \cite{zhou2022debiased}, CARD \cite{wang2022improving}. Although our method does not perform as good as CARD \cite{wang2022improving}, we argue that CARD is an orthogonal method in that it finetunes BERT/RoBERTa  with the help of finetuned models and additional data augmentation method,  and can be combined with IS-CSE.  We also include 7 transfer learning tasks \cite{conneau2017supervised}, taking STS as the main result for comparison following previous SimCSE-related papers \cite{gao2021simcse,wang2022improving,zhou2022debiased}. Our experients are conducted on one NVIDIA A100 GPU.

\subsection{Training Details}
Our experimental settings are consistent
with the SimCSE \cite{gao2021simcse}. Specifically, all our models are trained to start 
from the pre-trained checkpoints given by Huggingface \cite{wolf2020transformers}. Following SimCSE, the training corpus contains $10^6$ sentences randomly sampled from English
Wikipedia. We adopt [CLS] representation as the sentence embedding and an MLP pooler is used during training but discarded during inference. Hyperparameters for our model are the same as those for SimCSE. We train our model for 1 epoch and use the Adam optimizer \cite{kingma2014adam}. Cosine similarity with $\tau=0.05$ is used to calculate sentence similarity. The details of batch size and learning rate are shown in Table \ref{tab:lr_and_bs}.  In IS-CSE, we set the buffer size $L = 1024$ and the number of kNN neighbors $k=16$. According to the STS-B score on the development set in Table \ref{tab:knn_and_kmeans}, we finally select the kNN group to apply our smoothing method. The temperature $\beta$ for self-attention aggregation is set to 2. For BERT$_{base}$ and RoBERTa$_{base}$ we set $\alpha=0.1$. For BERT$_{large}$ and RoBERTa$_{large}$ we set a cosine schedule (Equ. \ref{equ: cosine_alpha}) for $\alpha$ from 0.005 to 0.05.

\begin{table}[tb]
\centering
\begin{tabular}{@{}lcccc@{}}
\toprule
\multicolumn{1}{c}{} & \multicolumn{2}{c}{BERT} & \multicolumn{2}{c}{RoBERTa} \\
\multicolumn{1}{c}{} & base       & large       & base         & large        \\ \midrule
Batch size           & 64         & 64          & 512          & 512          \\
Learning rate        & 3e-5       & 1e-5        & 1e-5         & 3e-5         \\ \bottomrule
\end{tabular}
\caption{Batch sizes and learning rates for IS-CSE}
\label{tab:lr_and_bs}
\end{table}
 
\begin{table}[tb]
\centering
\begin{tabular}{@{}lclccl@{}}
\toprule
Group type & \multicolumn{2}{c}{kNN}   & K-means & \multicolumn{2}{c}{kNN+K-means} \\ \midrule
STS-B                          & \multicolumn{2}{c}{\textbf{84.18}} & 83.74   & \multicolumn{2}{c}{84.14}       \\ \bottomrule
\end{tabular}
\caption{Results on STS-B development set of kNN group and K-means group using BERT$_{base}$ backbone.  For the K-means group, the number of groups is 64 so the average number of embeddings in each group is equal to that in the kNN group. kNN+K-means denotes that two groups will be used and thus two smoothing objectives will be added. kNN retrieval method is finally selected. }
\label{tab:knn_and_kmeans}
\end{table}

%%%%%%%%%%%%%%%%%%%%%%%%%%%%%%%%%%%%%%%%%%%%%%%%%%%%%%%%%%%
%%%%%%%%%%%%%%%%%%%%%%%%%%%%%%%%%%%%%%%%%%%%%%%%%%%%%%%%%%%
\subsection{Main Results}
We compare IS-CSE against previously published state-of-the-art unsupervised sentence embedding learning methods on STS tasks. We take the results reported in SimCSE for average GloVe embeddings \cite{pennington2014glove}, average BERT or RoBERTa embeddings \cite{gao2021simcse}, BERT-flow \cite{li2020sentence},  BERT-whitening \cite{su2021whitening}, unsup-SimCSE. For DCLR \cite{zhou2022debiased}, we take both the results reported on paper and our reproduced results based on their released code. 

The results on 7 STS tasks are shown in Table \ref{tabel: sts_main_resutls}. IS-CSE can outperform most previous competitive results on the basis of four different encoders (BERT$_{base}$, BERT$_{large}$, RoBERTa$_{base}$ and RoBERTa$_{large}$). Although we do not perform as well as DCLR on some of the tasks, IS-CSE is an orthogonal method in that it finetunes models with instance weighting, and may be combined with our methods. To evaluate it, we reproduce DCLR based on their released code and strictly follow their training settings. We further adding  $L_{smoothing}$ to DCLR and the results ( "+DCLR" in Table \ref{tabel: sts_main_resutls}) indicate that IS-CSE can improve DCLR on most STS tasks.

\subsection{Ablation Studies}
% \usepackage{booktabs}
We investigate the impact of buffer size $L$,  the hyper-parameter $\alpha, \beta$, and the number of neighbors in a group. All reported results in this section are based on the STS-B development set.

\subsubsection{Buffer Size}
Table \ref{table: the impact of buffer size} shows the results of IS-CSE-BERT$_{base}$ with different buffer sizes. As can be seen from Table \ref{table: the impact of buffer size}, when $L$ increases from 256 to 1024, the performance also improves, which shows that larger buffers can allow more similar instances to be retrieved. However, a large buffer size beyond 1024 may cause performance degradation, this can be because a large buffer stores embeddings of several batches, and older embeddings are inconsistent with the current model parameters.

%Table \ref{table: the impact of buffer size} shows IS-CSE-BERT$_{base}$ with a buffer length of 1024 outperforms others. Large buffer sizes can cause performance degradation, this can be because a large buffer can store embeddings of several batches, and older embeddings are inconsistent with the current model parameters. \textcolor{red}{hongliang: rewrite} %If the buffer size equals the size of the dataset, we retrieve sentences from the whole dataset. Table \ref{table: the impact of buffer size} shows that  

% {\color{red} hongliang xie}
\begin{table}[t]
\centering
\begin{tabular}{@{}lll@{}}
\toprule
Buffer size & STS-B & Avg. STS \\ \midrule
    256     & 82.75 &    78.95    \\
    512    & 83.85 &     79.48     \\
    1024    & \textbf{84.18} &    \textbf{79.91}      \\
    1536    & 83.02 &    78.35       \\
    2048    & 83.27 &    78.59      \\ 
    3072    & 83.60 &    78.92      \\ 
    \bottomrule
\end{tabular}
\caption{STS-B / Avg. STS development results with different buffer sizes using IS-CSE-BERT$_{base}$. }
\label{table: the impact of buffer size}
\end{table}

\subsubsection{Number of Neighbors} Table \ref{table: the impact of the number of neighbors} shows the effects of different numbers of neighbors in kNN. We empirically find that IS-CSE performs well when $k=16$, which is probably because that smoothing is not sufficient when $k$ is smaller than 16 and some noise samples will be introduced when $k$ is greater than 16.
%In kNN, we evaluate the influence of the number of neighbors. A small k value may make the smoothness insufficient and a large k value may introduce some noise samples. Table \ref{table: the impact of the number of neighbors} shows the effects of different  numbers of neighbors. \textcolor{red}{hongliang: rewrite}

\begin{table}[tb]
\centering
\begin{tabular}{@{}lllllll@{}}
\toprule
$N_{neighbors}$ &  8 & 12 & 16  & 20 &  24 & \\ \midrule
STS-B   &  83.18 &  83.31 & \textbf{84.18} & 82.97  & 82.63 & \\ \bottomrule
\end{tabular}
\caption{Ablation studies of the number of neighbors on the STS-B development set using IS-CSE-BERT$_{base}$. }
\label{table: the impact of the number of neighbors}
\end{table}

\subsubsection{Hyperparameter $\alpha$}
In IS-CSE, $\alpha$ is used as the weight of the $L_{smoothing}$. We tried two types of $\alpha$: constant  $\alpha$ and dynamic  $\alpha$. For the former, we just assign a constant value to $\alpha$ and never change it during finetuning; For the latter, we use a cosine schedule function (Equ. \ref{equ: cosine_alpha}) to gradually increase the value from $\alpha_{start}$ to $\alpha_{end}$. Table \ref{tab:impact_of_alpha} shows the result of applying different constant $\alpha$ and dynamic $\alpha$ on IS-CSE-RoBERTa$_{large}$. We empirically find that  BERT$_{large}$ and RoBERTa$_{large}$ can perform better with dynamic $\alpha$. 

\begin{table}[H]
\centering
\begin{tabular}{@{}ccccc@{}}
\toprule
 & $\alpha$ & $\alpha_{start}$ & $\alpha_{end}$ & STS-B \\ \midrule
 &  0.01 & -  & - &   85.12  \\
{Constant $\alpha$}&  0.05 & -  & - &   84.42  \\
&  0.1 & -  & - &   83.57  \\
  \midrule
  & - & 0.005  &      0.05  &   \textbf{85.76}  \\
 {Dynamic $\alpha$}& - & 0.05  &      0.1  &   84.47  \\
  & - & 0.005  &      0.1  &   84.99  \\
 \bottomrule
\end{tabular}
\caption{Effects of different $\alpha$ schedules on STS-B development set for IS-CSE-RoBERTa$_{large}$.}
\label{tab:impact_of_alpha}
\end{table}

\begin{table}[htb]
\centering
\begin{tabular}{@{}lllll@{}}
\toprule
$\beta$ &  1 & 2  &  3 & 4 \\ \midrule
STS-B    &  83.78 & \textbf{84.18} &  83.07 & 81.58 \\ \bottomrule
\end{tabular}
\caption{  Comparison of different constant $\beta$ on STS-B development set using IS-CSE-BERT$_{base}$. }
\label{table:the_impact_of_beta}
\end{table}

\subsubsection{Hyperparameter $\beta$}
After finishing the retrieval process, we perform self-attention aggregation on a group of embeddings to smooth the representation. In Table \ref{table:the_impact_of_beta}, we compare the impact of choosing different $\beta$ on STS-B development set. $\beta$ is used to adjust the attention weights, and a larger $\beta$ will make the attention weights more even.  %\textcolor{red}{hongliang: }

\begin{figure}
  \includegraphics[width=1.01\linewidth]{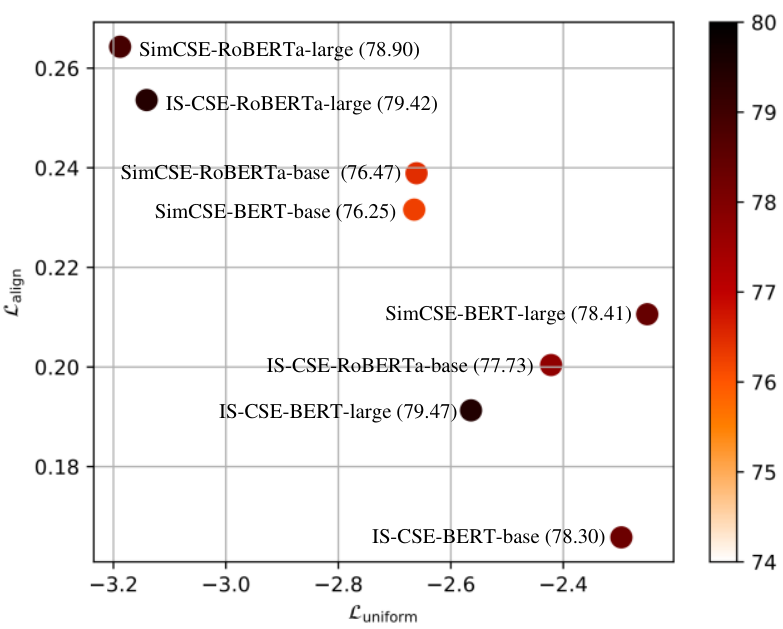}
  \caption{$\mathcal{L}_{align}$-$\mathcal{L}_{uniform}$ plot contains the alignment and uniformity measurements for our models. The color of points and numbers in brackets represent average STS performance. }
  \label{fig: uniformity and alignment}
\end{figure}

\begin{table*}[t]
\centering 
\begin{tabular}{@{}lllllllll@{}}
\toprule
Model                                 & MR    & CR    & SUBJ  & MPQA  & SST   & TREC  & MRPC  & Avg.  \\ \midrule
GloVe embeddings (avg.)               & 77.25 & 78.30 & 91.17 & 87.85 & 80.18 & 83.00 & 72.87 & 81.52 \\
Skip-thought                          & 76.50 & 80.10 & 93.60 & 87.10 & 82.00 & 92.20 & 73.00 & 83.50 \\
Avg. BERT embeddings                  & 78.66 & 86.25 & 94.37 & 88.66 & 84.40 & \textbf{92.80} & 69.54 & 84.94 \\
BERT-$\left [ CLS \right ]$ embedding & 78.68 & 84.85 & 94.21 & 88.23 & 84.13 & 91.40 & 71.13 & 84.66 \\
IS-BERT$_{base}$                      & 81.09 & \textbf{87.18} & \textbf{94.96} & 88.75 & \textbf{85.96} & 88.64 & 74.24 & \textbf{85.83} \\
SimCSE-BERT$_{base}$                  & \textbf{81.18} & 86.46 & 94.45 & 88.88 & 85.50 & 89.80 & 74.43 & 85.81 \\ \midrule
IS-CSE-BERT$_{base}$                  &  80.48	& 85.32	& 94.67 & \textbf{89.44} &	85.06 &	87.40 & \textbf{75.77} &	85.45   
    \\ \midrule
SimCSE-RoBERTa$_{base}$               & 81.04 & 87.74 & \textbf{93.28} & 86.94 & 86.60 & \textbf{84.60} & 73.68 & 84.84 \\
IS-CSE-RoBERTa$_{base}$                  & \textbf{81.93} & \textbf{87.76} &	93.24 &	\textbf{87.61} &	\textbf{87.48} &	83.20 &	\textbf{76.35} & 	\textbf{85.37}     \\
SimCSE-RoBERTa$_{large}$              & \textbf{82.74} & \textbf{87.87} & \textbf{93.66} & 88.22 & 88.58 & 92.00 & 69.68 & 86.11 \\

IS-CSE-RoBERTa$_{large}$                 & 82.70 &	87.79 &	93.30 &	\textbf{88.36} &	\textbf{89.02} &	\textbf{92.40} & 	\textbf{74.96} &	\textbf{86.93}  \\ 
\bottomrule
\end{tabular}
\caption{Transfer task results of different sentence embedding models (measured as accuracy). Results for comparison are reported in published paper SimCSE \cite{gao2021simcse}. We highlight the highest numbers among models with the same pre-trained encoder. } 
\label{tab: results of transfer learning}
\end{table*}
\begin{table*}[h!]
\centering
\begin{tabular}{@{}lll@{}}
\toprule
Query Sentence                     & \multicolumn{2}{l}{\begin{tabular}[c]{@{}l@{}}This can probably be attributed to the intelligence-gathering  of german civilians based \\ in ireland during the 1930s.\end{tabular}} \\ \midrule
\multirow{3}{*}{Retrieved Sentences} & 1          & \begin{tabular}[c]{@{}l@{}}The ``luftwaffe" carried out a number of air raids against the midlands and england \\ in the middle part of 1942.   \end{tabular}         \\
                                     & 2          & During the world war ii, the area became an important station for anti-activities                                                                            \\
                                     & 3          & Many union members were jewish and were killed during world war ii.                                                                                                  \\ \midrule
Query Sentence                      & \multicolumn{2}{l}{``ravenswood" may refer to}                                                                                                                                      \\ \midrule
\multirow{3}{*}{Retrieved Sentences} & 1          & ``roanoke" may refer to                                                                                                                                            \\
                                     & 2          & ``yasir ali" may refer to                                                                                                                                                \\
                                     & 3          &  ``datuna" may refer to                                                                                                                                                   \\ \bottomrule
\end{tabular}
\caption{We show the retrieved sentences in our method.  ``Query Sentence" represents the sentence used as a query.  ``Retrieved Sentences" represents the sentence retrieved from the dynamic memory buffer.}
\label{tab:re} 
\end{table*}

\subsection{Analysis}
In this section, we conduct further analyses to verify the effectiveness of IS-CSE.

\subsubsection{Alignment and Uniformity}
Alignment and Uniformity are two key properties of embedding learned by contrastive loss \cite{wang2020understanding}. It has been shown that models which have both better alignment and uniformity can perform better on sentence representation. 
% According to \cite{wang2020understanding}, if a distribution of positive pairs is given, the alignment loss can be measured by: 
% \begin{equation}
%     \label{equ:alignmentloss}
%     \mathcal{L}_{align} = 
% \end{equation}
Figure \ref{fig: uniformity and alignment} shows the uniformity and alignment of different sentence embedding models along with their STS results. Our smoothing method IS-CSE achieves better alignment on all four backbones.
However, compare to SimCSE, there are some adverse effects on the uniformity of base models. For large models, the uniformity only drops by 0.04 on RoBERTa$_{large}$ and increases by 0.31 on BERT$_{large}$. The results show that IS-CSE can achieve a better balance between alignment and uniformity. %\textcolor{red}{hongliang: this shows that..}

\begin{figure}[t!]
\centering
% \subfigure[IS-CSE] { \label{fig:a}
%  \includegraphics[width=0.5\columnwidth]{imgs/ISCSE.png}
% }
% \subfigure[SimCSE] { \label{fig:b}
% \includegraphics[width=0.5\columnwidth]{imgs/simcse.png}
% }
\subfigure[IS-CSE] { \label{fig:a}
    \begin{minipage}[t]{0.46\columnwidth}
        \centering
        \includegraphics[width=1.6in]{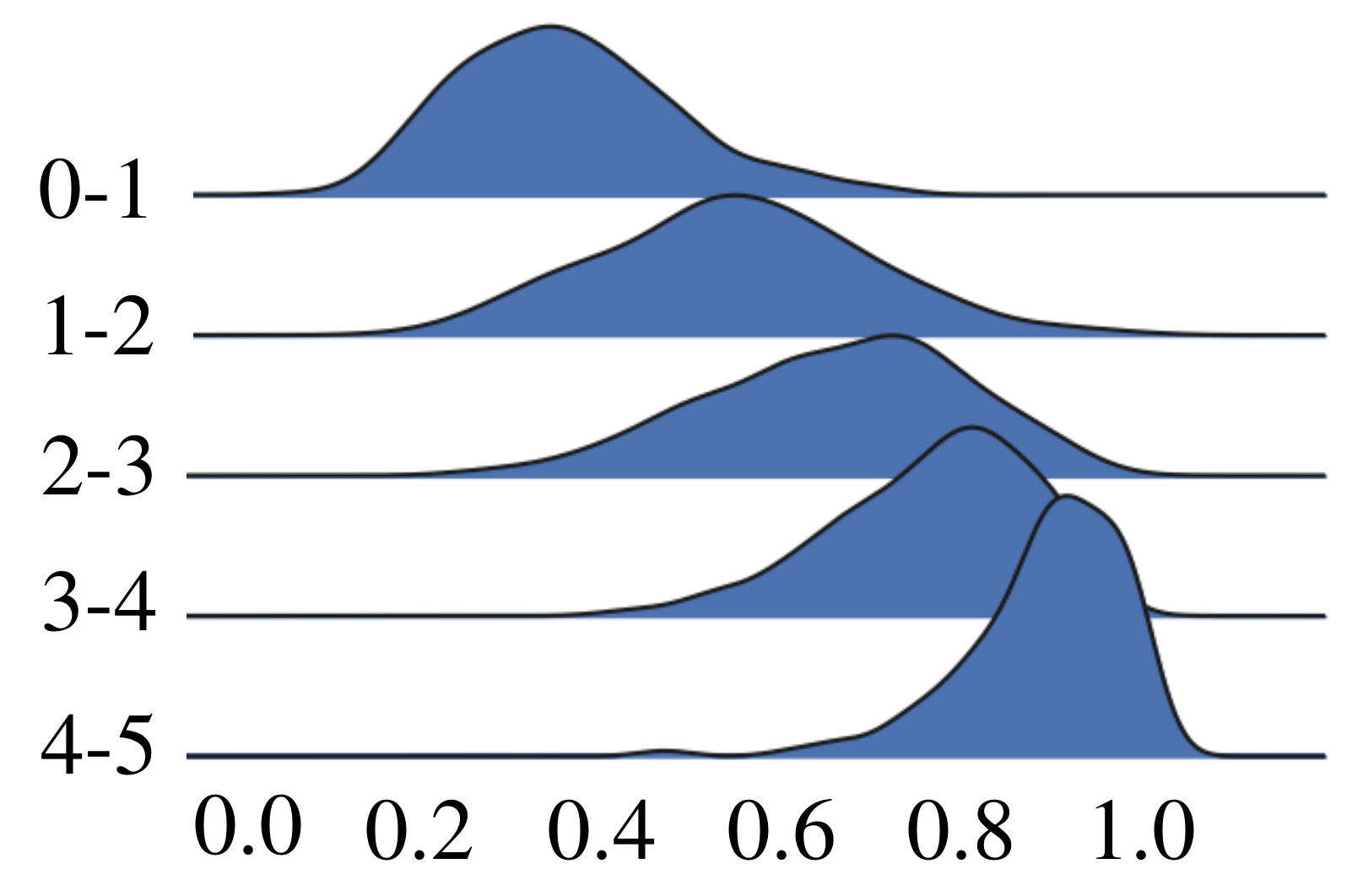}
    \end{minipage}
}
\subfigure[SimCSE] { \label{fig:b}
    \begin{minipage}[t]{0.46\columnwidth}
        \centering
        \includegraphics[width=1.6in]{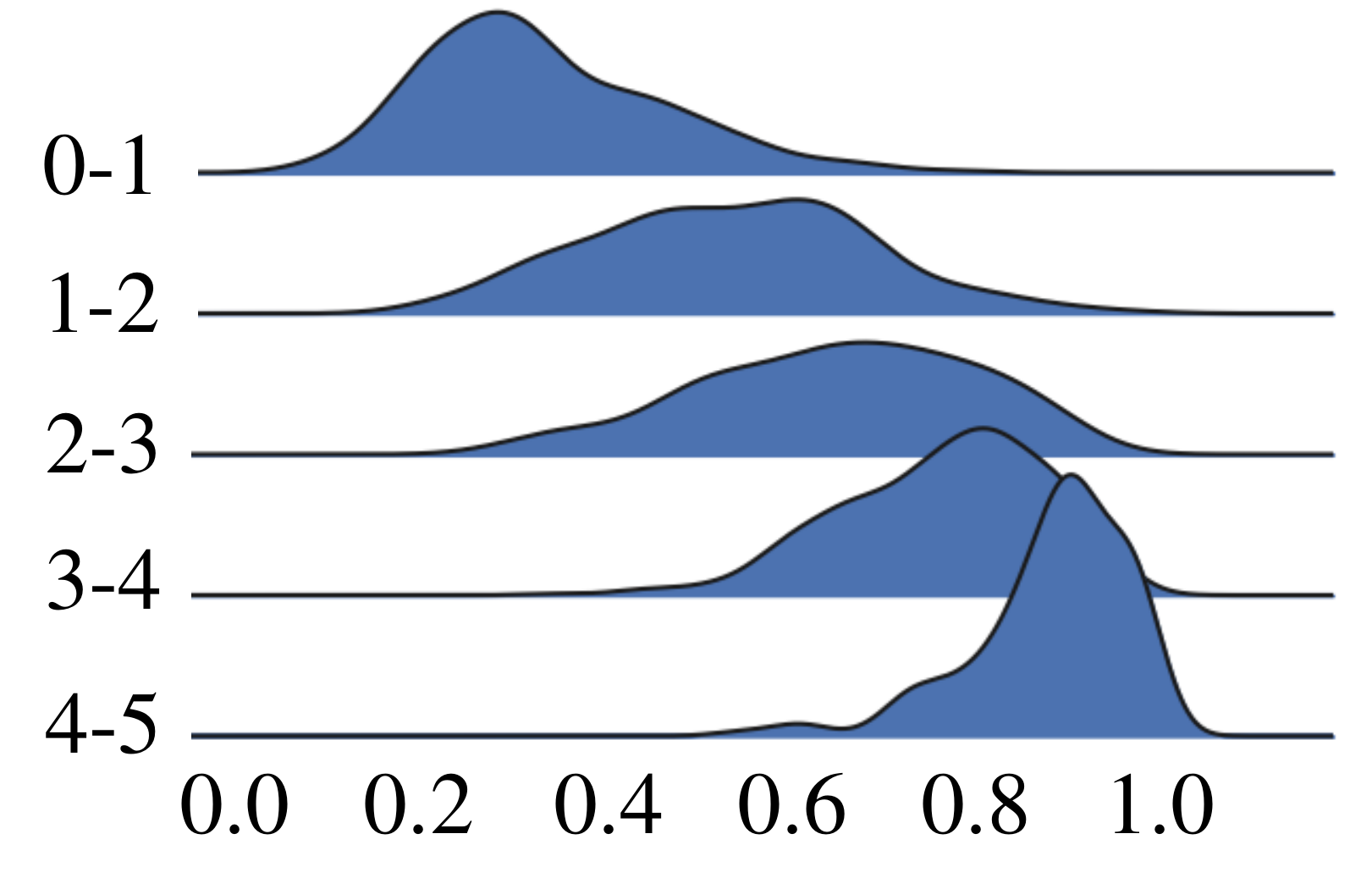}
    \end{minipage}
}
\caption{Density plots of cosine similarities between sentence pairs in STS-B based on the RoBERTa-large model. Sentence pairs are divided into 5 groups based on ground truth scores of similarity (higher means more similar) along the y-axis, and the x-axis is the cosine similarity.}

\label{fig:cosinesimilarity}
\end{figure}

\subsubsection{Transfer Learning}
We evaluate our models on the 7 transfer learning tasks: MR \cite{pang2005seeing}, CR \cite{hu2004mining}, MPQA \cite{wiebe2005annotating}, SST2 \cite{socher2013recursive}, TREC \cite{voorhees2000building} and MRPC \cite{dolan2005automatically}.  We train a logistic regression classifier on top of frozen finetuned encoders produced by different methods. We follow the default configurations in \cite{gao2021simcse}.

The results on transfer tasks are shown in Table \ref{tab: results of transfer learning}. IS-CSE achieves better or on par results than previous approaches except for BERT$_{base}$. This indicates that IS-CSE has better transferability than other models on most tasks. %\textcolor{red}{hongliang: transferability on most tasks} % SimCSE \cite{gao2021simcse} stated that transfer tasks are not a major goal for sentence embeddings. Following SimCSE, we take the STS results for the main comparison.

% \usepackage{booktabs}

\subsubsection{Results with Different Seeds}
\begin{figure}[t!]
\flushleft
  \includegraphics[width=0.99\linewidth,height=0.5\linewidth]{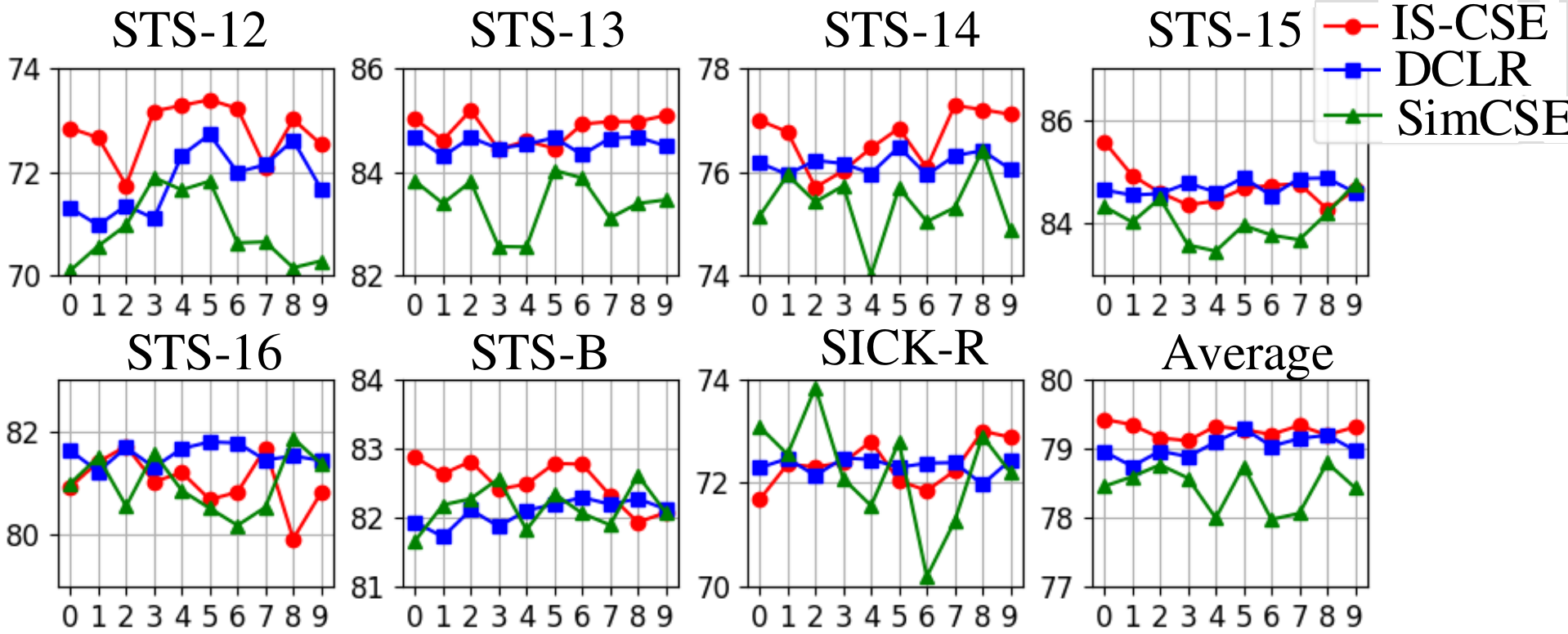}
  \caption{We compare our method on STS tasks with 10 random seeds based on RoBERTa-large model.}
  \label{fig: seedresults}
\end{figure}
To evaluate the stability of our method, we show the test result with four different seeds in Figure \ref{fig: seedresults}. Our model can outperform SimCSE and DCLR in most seeds and tasks.

\subsubsection{Cosine-Similarity Distribution}
To directly evaluate our approaches to STS tasks, we illustrate the cosine similarity distribution of sentence pairs in the STS-B dataset with different groups of human ratings in Figure \ref{fig:cosinesimilarity}. Compared with SimCSE, our method has a more scattered distribution with lower variance and has a similar discrimination ability. This observation validates that our method can achieve a better alignment-uniformity balance.

\subsubsection{Case Study of Retrieved Sentences}

We smooth the instance-level embeddings by aggregating retrieved embeddings. To better understand the smoothing process, we list the top three highest retrieved sentences based on kNN in Table \ref{tab:re}. The ``Query Sentence" is used as the query embedding during retrieval and the ``Retrieved Sentences" are the top three highest sentences retrieved from the dynamic memory buffer according to the similarity. Though the meaning of retrieved sentences and the query sentences is not totally the same, they are similar semantically in some text segments. For example, the query sentence ``ravenswood may refer to" has the same structure as retrieved sentence ``roanoke” may refer to". Thus the retrieved sentences help to smooth the query sentence and achieve better performance on STS tasks. % \textcolor{red}{example, similar sentence structure}

\section{Conclusion}
We proposed IS-CSE, an instance smoothing contrastive learning framework for unsupervised sentence representation learning. Our main idea is to improve the generalization ability by smoothing the positive examples. Specifically, in our framework, we aggregate retrieved semantically similar instances from a dynamic memory buffer to produce group-level positive embeddings, which are then used for discrimination. Experimental results on seven STS tasks have shown that our approach outperforms several competitive baselines. Our instance-level smoothing method is general and can be applied to other settings in Contrastive Learning.

In the future, we will explore more granularities for smoothing positive sentences for discrimination. Whether negative examples can be smoothed will also be studied. We will also consider applying our method for more natural language processing tasks, such as summarization. 

\section{Acknowledgements}
This research has been supported by the Key R\&D program of Zhejiang Province (Grant No. 2021C03139). We also would like to thank Westlake University HPC Center for providing HPC support.
%%%%%%%%%%%%%%%%%%%%%%%%%%%%%%%%%%%%%%%%%%%%%%%%%%%%%%%%%

\bibliography{aaai23.bib}

\end{document}